\documentclass{article}
\usepackage[utf8]{inputenc}
\usepackage{authblk}
\usepackage{setspace}
\usepackage[margin=1.25in]{geometry}
\usepackage{graphicx}
\usepackage{booktabs}
\usepackage{caption}
\usepackage{subcaption}
\graphicspath{ {./figures/} }
\usepackage{subcaption}
\usepackage{amsmath}
\usepackage[colorlinks = true]{hyperref}
\usepackage[noadjust]{cite}
\usepackage{pgf}
\usepackage[absolute, overlay]{textpos}

\textblockorigin{\paperwidth}{0.5 pt}

\newcommand\blfootnote[1]{%
  \begingroup
  \renewcommand\thefootnote{}\footnote{#1}%
  \addtocounter{footnote}{-1}%
  \endgroup
}

\title{Performance Analysis of Support Vector Machine (SVM) on Challenging Datasets for Forest Fire Detection}

\author[1,a]{Ankan Kar}
\author[2,b]{Nirjhar Nath}
\author[3,c]{Utpalraj Kemprai}
\author[3,d]{Aman}

\affil[1]{Department of Computer Science, Chennai Mathematical Institute}
\affil[2]{Department of Mathematics and Computer Science, Chennai Mathematical Institute}
\affil[3]{Department of Data Science, Chennai Mathematical Institute}

\date{December 2023}

\begin{document}

\maketitle

\pgfmathwidth{"top right corner at page \thepage"}
%% the width of text is stored in '\pgfmathresult'
\begin{textblock}{\pgfmathresult}[1, 0](0, 0)
%% value of '\pgfmathresult' is used to set the width of text block
%% '[1, 0]' sets the anchor point of text block to be its top right corner
%% '(0, 0)' sets the anchor point right at the origin which is set by '\textblockorigin' in the preamble
\noindent
\large
\textbf{International Journal of Communications, Network and System Sciences, 17, 11-29}. \\ DOI: \href{https://doi.org/10.4236/ijcns.2024.172002}{10.4236/ijcns.2024.172002}\\
\textbf{Publication Date}: February 29, 2024
\end{textblock}

%%%%%% Abstract %%%%%%
\begin{abstract}
This article delves into the analysis of performance and utilization of Support Vector Machines (SVMs) for the critical task of forest fire detection using image datasets. With the increasing threat of forest fires to ecosystems and human settlements, the need for rapid and accurate detection systems is of utmost importance. SVMs, renowned for their strong classification capabilities, exhibit proficiency in recognizing patterns associated with fire within images. By training on labeled data, SVMs acquire the ability to identify distinctive attributes associated with fire, such as flames, smoke, or alterations in the visual characteristics of the forest area. The document thoroughly examines the use of SVMs, covering crucial elements like data preprocessing, feature extraction, and model training. It rigorously evaluates parameters such as accuracy, efficiency, and practical applicability. The knowledge gained from this study aids in the development of efficient forest fire detection systems, enabling prompt responses and improving disaster management. Moreover, the correlation between SVM accuracy and the difficulties presented by high-dimensional datasets is carefully investigated, demonstrated through a revealing case study. The relationship between accuracy scores and the different resolutions used for resizing the training datasets has also been discussed in this article. These comprehensive studies result in a definitive overview of the difficulties faced and the potential sectors requiring further improvement and focus.\\

\textbf{Keywords:} Support Vector Machine, Challenging Datasets, Forest Fire Detection, Classification
\end{abstract}

%%%%%% Main Text %%%%%%

\blfootnote{$^\text{a}$\href{mailto:zargon.ankan@gmail.com}{zargon.ankan@gmail.com},
$^\text{b}$\href{mailto:nirjharnath31595@gmail.com}{nirjharnath31595@gmail.com}, $^\text{c}$\href{mailto:utpalrajkemprai2001@gmail.com}{utpalrajkemprai2001@gmail.com}, $^\text{d}$\href{mailto:amanbhoot84@gmail.com}{amanbhoot84@gmail.com}}

\section*{Introduction}
Support Vector Machines (SVMs) represent a powerful class of supervised machine learning algorithms renowned for their versatility and effectiveness in solving a wide range of classification and regression tasks. Introduced by Vladimir Vapnik and his colleagues in the 1960s, SVMs have since become a cornerstone of modern machine learning.

At their core, SVMs excel in finding optimal decision boundaries that separate data points belonging to different classes while maximizing the margin, or distance, between these boundaries. This unique characteristic allows SVMs to perform exceptionally well in scenarios where data may be complex, high-dimensional, or not linearly separable. Moreover, SVMs are known for their ability to generalize from training data to new, unseen examples, which makes them valuable tools for both classification and regression problems. We can see a similar work in \cite{Mohan} and \cite{Chidambaram}. A comparative analysis of forest fire detection can be seen in \cite{Susmitha}.

High-dimensional datasets refer to datasets where each data point has a large number of features or dimensions. A dataset can be represented as a matrix, where each row corresponds to a data point, and each column corresponds to a feature. High-dimensional datasets have a large number of columns or dimensions, making them challenging to visualize and analyze. Collecting, storing, and processing this extensive information may not contribute additional advantages to optimal decision-making; instead, it could potentially complicate matters and incur excessive costs. In \cite{Ghaddar}, Ghaddar et al. address the problem of feature selection within SVM classification that deals with finding an accurate binary classifier that uses a minimal number of features available in the high-dimensional datasets.

SVMs have found applications in diverse fields, including image classification, text categorization, biological sciences, finance, and more. Their adaptability, robustness, and capacity to handle large datasets make them a preferred choice for many researchers and practitioners in the domain of machine learning.

%%%%% Citations in the text %%%%%%
% \subsection*{Citations}
% Citations of references in the text should be identified using numbers in square brackets e.g., ``as discussed by Cui \cite{Cui1}'' or ``as discussed elsewhere \cite{Cui1,Ninomiya1,Li1,Wang1,Yang1}.'' All references should be cited within the text and uncited references will be removed. 

% As an example, this template includes a ``sample.bib'' file containing the references in BibTeX.

%%%%%% Optimization in Machine Learning %%%%%%
\section*{Optimization in Machine Learning}
Machine learning tasks often follow a common structure: we start with a dataset containing pairs of input and output, like $\{(x_1, y_1), (x_2, y_2), ..., (x_n, y_n)\}$. The goal is to find a function, denoted as $f_\theta$, where $\theta$ represents a set of parameters from a predefined set $P$. We want this function to minimize a loss function, which measures the difference between the predicted output $f_\theta(x)$ and the actual output $y$.

So, in a formal sense, machine learning tasks can be boiled down to two optimization problems:

1. Optimization in the space of possible functions $M$, where we seek to minimize the loss:
   \[
   \min_{f_\theta \in M} l(y, f_\theta(x))
   \]

2. Optimization in the space of parameters $P$, where we aim to minimize the loss by adjusting $\theta$:
\[\min_{\theta \in P} l(y, f_\theta(x))\]

To prevent overfitting and enhance the model's ability to generalize to new data, we may include regularization terms in the loss function. This leads to a modified optimization problem:
\[ \min_{\theta \in P} \left(l(y, f_\theta(x)) + \lambda R(\theta)\right) \]

Here, $R(\theta)$ represents the regularization term, and $\lambda$ is a hyperparameter controlling the strength of regularization.
The details can be found in \cite{Vahid} and \cite{Gilles}.

%%%%%% From Linear to Non-linear via Kernel Method %%%%%%
\subsection*{Kernel Used For Optimization}
The kernel method represents a prominent machine learning technique, specifically designed to address nonlinear classification and regression problems. In numerous practical scenarios, the association between input variables and the target variable does not stick to linear patterns. In such instances, conventional linear models like linear regression or logistic regression may exhibit suboptimal performance. (In \cite{Bernhard}, Sch\"{o}lkopf et al. provides an introduction to SVMs and related kernel methods including the latest research in this area.)

The kernel method provides a remedy by facilitating the transformation of input variables into a higher-dimensional feature space. Within this feature space, it becomes possible to establish linear relationships between the input variables and the target variable. This transformation is facilitated through the utilization of a kernel function, which is a mathematical function designed to quantify the similarity between pairs of input data points.

By projecting the input data into this higher-dimensional feature space, the kernel method is capable of capturing intricate and nonlinear relationships between the input variables and the target variable. It also offers the versatility to complement various machine learning algorithms, including support vector machines (SVMs), ridge regression, and principal component analysis (PCA).

From a mathematical perspective, the kernel method involves the mapping of input data points, represented as vectors denoted as \(x\) within a \(d\)-dimensional input space, into a higher-dimensional feature space denoted as \(F\), achieved through the application of a kernel function labeled as \(K\).

This kernel function operates by taking a pair of input vectors, namely \(x\) and \(x'\), and subsequently computing the dot product of their corresponding feature representations within \(F\). Formally, the kernel function can be articulated as follows: \(K(x, x') = \langle\phi(x), \phi(x')\rangle\), where \(\phi(x)\) and \(\phi(x')\) respectively denote the feature representations of \(x\) and \(x'\) within the feature space \(F\).

The selection of an appropriate kernel function is a pivotal decision in machine learning, contingent on the specific problem at hand and the unique characteristics of the input data. Several common kernel functions serve distinct purposes:

\begin{itemize}
  \item \textbf{Linear Kernel:} This kernel, denoted as $K(x, x') = x^T x'$, assumes a linear relationship between data points. It is suitable when the underlying data patterns follow linear trends.

  \item \textbf{Polynomial Kernel:} The polynomial kernel, defined as $K(x, x') = (x^T x' + c)^d$, introduces nonlinearity by elevating the dot product of data points to a certain power ($d$), while $c$ represents a constant. This kernel is versatile and can effectively capture more intricate relationships within the data.

  \item \textbf{Gaussian (RBF) Kernel:} The Gaussian or Radial Basis Function (RBF) kernel, expressed as $K(x, x') = \exp\left(-\frac{||x - x'||^2}{2\sigma^2}\right)$, relies on the Euclidean distance between data points. The parameter $\sigma$ governs the width of the Gaussian function, enabling adaptation to data with varying scales. It excels at capturing intricate patterns in the data.

  \item \textbf{Sigmoid Kernel} The sigmoid kernel in machine learning is a similarity measure used for classification tasks. It is defined as $K(x, x') = \tanh(\alpha x^T x' + \beta)$,
  where \(\alpha\) and \(\beta\) are hyperparameters controlling the kernel's shape. It captures non-linear relationships between data points and is particularly useful when data has complex, sigmoid-shaped decision boundaries.

\end{itemize}

Once the input data has been transformed into the higher-dimensional feature space $F$ using a selected kernel function, linear algorithms such as Support Vector Machines (SVMs) or ridge regression can be applied for classification or regression tasks. Within this transformed feature space, the decision boundary is represented as a hyperplane, often corresponding to a nonlinear decision boundary when projected back into the original input space. This approach empowers the handling of complex, nonlinear relationships present in the data.

In practical machine learning applications, the feature representations denoted as $\phi(x)$ within the higher-dimensional feature space $F$ are frequently not explicitly calculated. This is made possible by the kernel function, which enables the computation of dot products between data points without the need to construct and store the feature vectors themselves. This remarkable technique, often referred to as the ``kernel trick", bestows upon the kernel method the computational efficiency necessary to operate effectively within high-dimensional feature spaces. As a result, it simplifies the procedure of managing intricate data alterations while preserving computational feasibility. The details can be found in \cite{Vahid}.

%%%%%% Data Availability %%%%%%
\section*{Data Availability}
One of the dataset employed for this study is conveniently accessible in \href{https://data.mendeley.com/datasets/gjmr63rz2r/1}{Dataset for Forest Fire Detection in Mendeley Data}. It is provided as a compressed file ``Dataset.rar". This archive contains two essential components: the training dataset and the test dataset. These datasets consist of images, each having a resolution of 250$\times$250 pixels.

\begin{figure}[h!]
     \centering
     \begin{subfigure}[b]{0.147\textwidth}
         \centering
         \includegraphics[width=\textwidth]{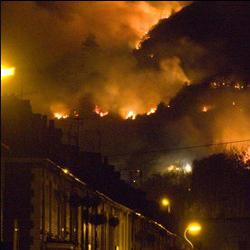}
         \caption{}
         \label{}
     \end{subfigure}
     \hfill
     \begin{subfigure}[b]{0.147\textwidth}
         \centering
         \includegraphics[width=\textwidth]{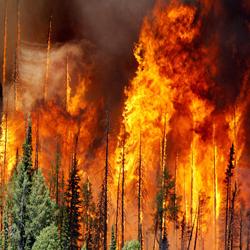}
         \caption{}
         \label{}
     \end{subfigure}
     \hfill
     \begin{subfigure}[b]{0.147\textwidth}
         \centering
         \includegraphics[width=\textwidth]{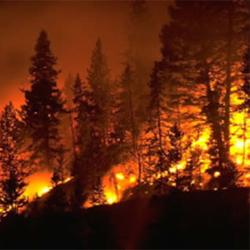}
         \caption{}
         \label{}
     \end{subfigure}
     \hfill
     \begin{subfigure}[b]{0.147\textwidth}
         \centering
         \includegraphics[width=\textwidth]{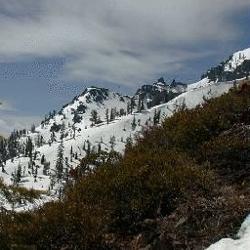}
         \caption{}
         \label{}
     \end{subfigure}
     \hfill
     \begin{subfigure}[b]{0.147\textwidth}
         \centering
         \includegraphics[width=\textwidth]{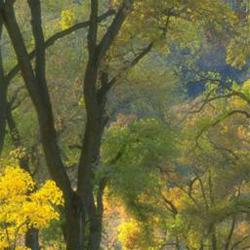}
         \caption{}
         \label{}
     \end{subfigure}
     \hfill
     \begin{subfigure}[b]{0.147\textwidth}
         \centering
         \includegraphics[width=\textwidth]{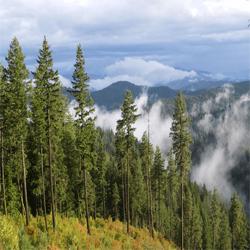}
         \caption{}
         \label{}
     \end{subfigure}
     \centering
     \begin{subfigure}[b]{0.147\textwidth}
         \centering
         \includegraphics[width=\textwidth]{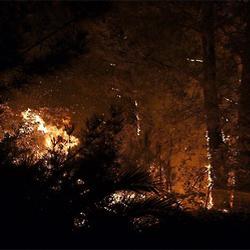}
         \caption{}
         \label{}
     \end{subfigure}
     \hfill
     \begin{subfigure}[b]{0.147\textwidth}
         \centering
         \includegraphics[width=\textwidth]{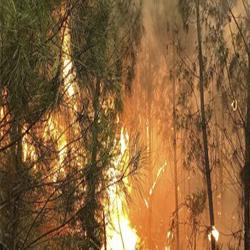}
         \caption{}
         \label{}
     \end{subfigure}
     \hfill
     \begin{subfigure}[b]{0.147\textwidth}
         \centering
         \includegraphics[width=\textwidth]{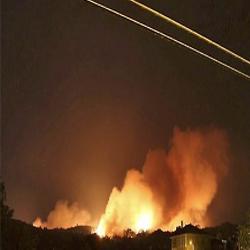}
         \caption{}
         \label{}
     \end{subfigure}
     \hfill
     \begin{subfigure}[b]{0.147\textwidth}
         \centering
         \includegraphics[width=\textwidth]{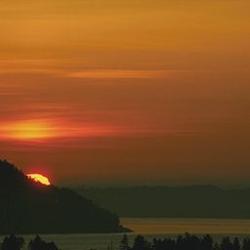}
         \caption{}
         \label{}
     \end{subfigure}
     \hfill
     \begin{subfigure}[b]{0.147\textwidth}
         \centering
         \includegraphics[width=\textwidth]{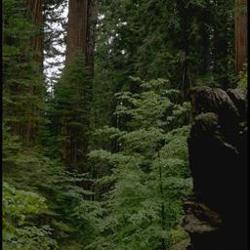}
         \caption{}
         \label{}
     \end{subfigure}
     \hfill
     \begin{subfigure}[b]{0.147\textwidth}
         \centering
         \includegraphics[width=\textwidth]{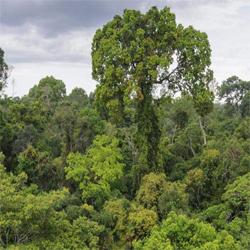}
         \caption{}
         \label{}
     \end{subfigure}
     \hfill
     \caption{Images of the upper and the lower row belong to the training and testing datasets respectively of ``Dataset.rar" with: (a),(b),(c) Elements of the training dataset with fire (d),(e),(f) Elements of the training dataset with no fire (g),(h),(i) Elements of the test dataset with fire (j),(k),(l) Elements of the test dataset with no fire.}
\end{figure}
The other dataset we used for checking the efficiency of our model is  accessible from \href{https://images.cv/download/forest_fire/948/CALL_FROM_SEARCH/%22forest_fire%22}{\text{images.cv}}. In this dataset all images are of 256$\times$256 pixels and all belong to the category of forest\_fire. They have been curated to focus specifically on imagery related to forest fires. 

\begin{figure}[h!]
     \centering
     \begin{subfigure}[b]{0.147\textwidth}
         \centering
         \includegraphics[width=\textwidth]{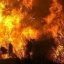}
         \caption{}
         \label{}
     \end{subfigure}
     \hfill
     \begin{subfigure}[b]{0.147\textwidth}
         \centering
         \includegraphics[width=\textwidth]{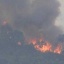}
         \caption{}
         \label{}
     \end{subfigure}
     \hfill
     \begin{subfigure}[b]{0.147\textwidth}
         \centering
         \includegraphics[width=\textwidth]{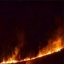}
         \caption{}
         \label{}
     \end{subfigure}
     \hfill
     \begin{subfigure}[b]{0.147\textwidth}
         \centering
         \includegraphics[width=\textwidth]{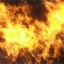}
         \caption{}
         \label{}
     \end{subfigure}
     \hfill
     \begin{subfigure}[b]{0.147\textwidth}
         \centering
         \includegraphics[width=\textwidth]{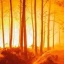}
         \caption{}
         \label{}
     \end{subfigure}
     \hfill
     \begin{subfigure}[b]{0.147\textwidth}
         \centering
         \includegraphics[width=\textwidth]{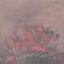}
         \caption{}
         \label{}
     \end{subfigure}
     \centering
     \begin{subfigure}[b]{0.147\textwidth}
         \centering
         \includegraphics[width=\textwidth]{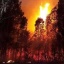}
         \caption{}
         \label{}
     \end{subfigure}
     \hfill
     \begin{subfigure}[b]{0.147\textwidth}
         \centering
         \includegraphics[width=\textwidth]{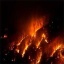}
         \caption{}
         \label{}
     \end{subfigure}
     \hfill
     \begin{subfigure}[b]{0.147\textwidth}
         \centering
         \includegraphics[width=\textwidth]{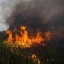}
         \caption{}
         \label{}
     \end{subfigure}
     \hfill
     \begin{subfigure}[b]{0.147\textwidth}
         \centering
         \includegraphics[width=\textwidth]{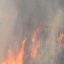}
         \caption{}
         \label{}
     \end{subfigure}
     \hfill
     \begin{subfigure}[b]{0.147\textwidth}
         \centering
         \includegraphics[width=\textwidth]{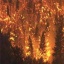}
         \caption{}
         \label{}
     \end{subfigure}
     \hfill
     \begin{subfigure}[b]{0.147\textwidth}
         \centering
         \includegraphics[width=\textwidth]{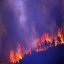}
         \caption{}
         \label{}
     \end{subfigure}
     \hfill
     \caption{Images of the upper and the lower row belong to the training and testing datasets of images.cv respectively.}
\end{figure}

%%%%%% Aim %%%%%%
\section*{Aim}
The primary focus of this report is to thoroughly analyze the performance of Support Vector Machines (SVMs) in the context of forest fire detection, with a particular emphasis on challenging datasets characterized by high dimensionality and limited samples. A unified theory for general class of nonconvex penalized SVMs
in the high-dimensional setting can be found in \cite{Zhang}. In our article, the overarching objective is to contribute to the development of a model for enhancing the efficiency of forest fire detection using image data.

Specifically, our aim is to examine the performance of SVMs under challenging conditions, understanding how the model responds to high-dimensional datasets with sparse samples. The goal is to gain insights into the variations in performance and to identify the factors that influence the model's accuracy in detection. Take a look at \cite{Sobha} for a quick idea in the field of machine learning in Forest Fire Detection. An empirical study on the viability of SVM in the context of feature selection from moderately and highly unbalanced datasets can be seen in \cite{Chinna}.

Through this analysis, we strive to acquire a thorough comprehension of the model’s reactions under various circumstances. This understanding is crucial in pinpointing opportunities for growth and enhancement that positively influence the model, ultimately resulting in improved precision in forest fire detection.

%%%%%% Implementation %%%%%%
\section*{Implementation}
Our dataset consists of labeled images categorized as ``fire" and ``no-fire." The primary objective of this study was to employ predictive methods to determine whether unseen test data belonged to either the ``fire" or ``no-fire" category. Given the binary nature of this classification task, several techniques were at our disposal. Ultimately, we opted to train a Support Vector Machine (SVM) model for this purpose. This SVM model was carefully trained to classify images and predict the occurrence of forest fires based on the visual content of the images. This choice was made after careful consideration of its suitability for binary classification tasks involving image data. While there were options to employ diverse dimension reduction methods for handling high-dimensional datasets, our focus was to assess the performance of the SVM model under these conditions. Consequently, we refrained from engaging in dimension reduction techniques. However, dimensionality reduction techniques such as Principal Component Analysis, Linear Discriminant Analysis, Independent Component Analysis, Canonical Correlation Analysis, Fisher’s Linear Discriminant, Topic Models and Latent Dirichlet
Allocation, etc. could have been used to increase the efficiency of the SVM model.

\subsection*{Procedure}
Our dataset presents a challenge in which the number of data samples is notably low in comparison to the multitude of attributes considered. Specifically, we have utilized the pixel values of images as these attributes. To address this issue, we undertook a series of preprocessing steps to enhance the dataset's suitability for analysis.

First, we performed resizing of the images to various resolutions, including: $10\times10, 20\times20, 30\times30, 40\times40, 50\times50, 60\times60, 70\times70, 80\times80, 90\times90, 100\times100, 150\times150, 200\times200, 250\times250$. This resizing was performed exclusively on the training datasets to better comprehend the relationship between the quantity of samples and the number of attributes.

Furthermore, in our desire for increasing the number of data samples, we applied data augmentation techniques like flip and median blur which resulted in increase of 4 times the samples we have - original samples, flipped version of original samples and median blurred samples of both original and flipped samples. These augmentations were essential in the development of a more robust dataset for subsequent analysis.

Subsequently, we explored classification methodologies, utilizing Support Vector Machines (SVM) with polynomial, sigmoid, and Gaussian kernels. Additionally, we employed 4$-$Folds Cross Validation and the Grid Search algorithm on the various resized image datasets to assess their classification performance. Furthermore, we conducted a comparative study of the SVM models under these challenging conditions by applying Logistic Regression to the same image datasets. The samples were taken as input of SVM as vectors of values of red, green and blue values of all the pixels of the images. The SVM model is then run on various values of parameters for kernels and among them the best model is chosen for observations.

It is important to note that the resizing and data augmentation procedures were exclusively applied to the training datasets. The two test datasets remained unaltered and were treated as unseen data during our model evaluation, thereby ensuring a robust assessment of model generalization and performance. Our model was quite simple as we are only interested in the performance. Some prediction related works can be seen as in \cite{Eashwar} and \cite{Yanjie}.

\subsection*{Results \& Observations}
In the thorough assessment of our models carried out on both balanced and unbalanced datasets, it is apparent that specific classifiers display different levels of performance, providing valuable understanding into their effectiveness within the scope of the given classification task. This examination aims to clarify the relative advantages and disadvantages of these classifiers, thereby giving a comprehensive comprehension of their performance.

\begin{table}[!b]
\centering
\caption{Logistic Regression on Balanced Dataset}
\label{tab:logistic-regression-results}
\begin{tabular}{ccc}
\toprule
Resolution Size & Accuracy Score & Confusion Matrix \\
\midrule
10   & 0.868421 & $[[165, 25], [25, 165]]$ \\
20   & 0.852632 & $[[163, 27], [29, 161]]$ \\
30   & 0.834211 & $[[159, 31], [32, 158]]$ \\
40   & 0.813158 & $[[153, 37], [34, 156]]$ \\
50   & 0.834211 & $[[155, 35], [28, 162]]$ \\
60   & 0.865789 & $[[160, 30], [21, 169]]$ \\
70   & 0.863158 & $[[164, 26], [26, 164]]$ \\
80   & 0.868421 & $[[163, 27], [23, 167]]$ \\
90   & 0.873684 & $[[166, 24], [24, 166]]$ \\
100  & 0.871053 & $[[164, 26], [23, 167]]$ \\
150  & 0.873684 & $[[166, 24], [24, 166]]$ \\
200  & 0.868421 & $[[164, 26], [24, 166]]$ \\
250  & 0.868421 & $[[163, 27], [23, 167]]$ \\
\bottomrule
\end{tabular}
\end{table}

\begin{table}[!b]
\centering
\caption{Logistic Regression on Unbalanced Dataset}
\label{tab:logistic-regression-unbalanced-results}
\begin{tabular}{cccc}
\toprule
Resolution Size & Accuracy Score & Confusion Matrix \\
\midrule
10   & 0.947020 & $[[0, 0], [48, 858]]$ \\
20   & 0.964680 & $[[0, 0], [32, 874]]$ \\
30   & 0.962472 & $[[0, 0], [34, 872]]$ \\
40   & 0.965784 & $[[0, 0], [31, 875]]$ \\
50   & 0.973510 & $[[0, 0], [24, 882]]$ \\
60   & 0.976821 & $[[0, 0], [21, 885]]$ \\
70   & 0.972406 & $[[0, 0], [25, 881]]$ \\
80   & 0.974614 & $[[0, 0], [23, 883]]$ \\
90   & 0.973510 & $[[0, 0], [24, 882]]$ \\
100  & 0.976821 & $[[0, 0], [21, 885]]$ \\
150  & 0.975717 & $[[0, 0], [22, 884]]$ \\
200  & 0.974614 & $[[0, 0], [23, 883]]$ \\
250  & 0.975717 & $[[0, 0], [22, 884]]$ \\
\bottomrule
\end{tabular}
\end{table}

\begin{table}[t]
\centering
\caption{Sigmoid Kernel SVM on Balanced Dataset}
\label{tab:sigmoid-svm-balanced-results}
\begin{tabular}{ccc}
\toprule
Resolution Size & Accuracy Score & Confusion Matrix \\
\midrule
10   & 0.473684 & $[[81, 109], [91, 99]]$ \\
20   & 0.689474 & $[[142, 48], [70, 120]]$ \\
30   & 0.689474 & $[[142, 48], [70, 120]]$ \\
40   & 0.689474 & $[[142, 48], [70, 120]]$ \\
50   & 0.689474 & $[[142, 48], [70, 120]]$ \\
60   & 0.689474 & $[[142, 48], [70, 120]]$ \\
70   & 0.689474 & $[[141, 49], [69, 121]]$ \\
80   & 0.689474 & $[[141, 49], [69, 121]]$ \\
90   & 0.689474 & $[[141, 49], [69, 121]]$ \\
100  & 0.689474 & $[[141, 49], [69, 121]]$ \\
150  & 0.689474 & $[[141, 49], [69, 121]]$ \\
200  & 0.507895 & $[[90, 100], [87, 103]]$ \\
250  & 0.689474 & $[[141, 49], [69, 121]]$ \\
\bottomrule
\end{tabular}
\end{table}

\begin{table}[t]
\centering
\caption{Sigmoid Kernel SVM on Unbalanced Dataset}
\label{tab:sigmoid-svm-unbalanced-results}
\begin{tabular}{ccc}
\toprule
Resolution Size & Accuracy Score & Confusion Matrix \\
\midrule
10   & 0.539735 & $[[0, 0], [417, 489]]$ \\
20   & 0.663355 & $[[0, 0], [305, 601]]$ \\
30   & 0.663355 & $[[0, 0], [305, 601]]$ \\
40   & 0.666667 & $[[0, 0], [302, 604]]$ \\
50   & 0.667770 & $[[0, 0], [301, 605]]$ \\
60   & 0.668874 & $[[0, 0], [300, 606]]$ \\
70   & 0.671082 & $[[0, 0], [298, 608]]$ \\
80   & 0.671082 & $[[0, 0], [298, 608]]$ \\
90   & 0.671082 & $[[0, 0], [298, 608]]$ \\
100  & 0.671082 & $[[0, 0], [298, 608]]$ \\
150  & 0.673289 & $[[0, 0], [296, 610]]$ \\
200  & 0.556291 & $[[0, 0], [402, 504]]$ \\
250  & 0.673289 & $[[0, 0], [296, 610]]$ \\
\bottomrule
\end{tabular}
\end{table}

First and foremost, the Sigmoid Kernel Support Vector Machine (SVM) has emerged as the weakest contender within the spectrum of classifiers examined. Its performance was observed to be suboptimal, failing to meet the standards set by other classifiers. Significantly, when compared to the Logistic Regression model, the Sigmoid Kernel SVM demonstrated a noticeably poorer performance, thereby highlighting its inappropriateness for the specific classification task being examined. It is evident that this particular SVM variant struggled to recognize and categorize patterns within the data effectively, leading to a comparatively higher rate of misclassification.

In stark contrast, the Polynomial Kernel SVM showcased a more promising performance trajectory. In direct comparison to the Logistic Regression model, the Polynomial Kernel SVM managed to outperform the latter. This indicates that the former possesses a certain degree of resilience and robustness in handling the complications of the dataset. It is worth noting that the Polynomial Kernel SVM, with its capacity to model complex, nonlinear relationships, demonstrated an inherent advantage over the logistic regression model, which tends to assume linearity in its decision boundaries.

Further refinement in the classification outcomes was observed with the Gaussian Kernel SVM. This specific variant of the Support Vector Machine displayed superior performance when compared to all the classifiers discussed in this evaluation. Its enhanced effectiveness in identifying and categorizing instances can be credited to the Gaussian kernel's capacity to grasp complex patterns and nonlinearity, which might be present in the dataset. The Gaussian Kernel SVM, therefore, presents itself as a formidable choice when intricate and nonlinear relationships are inherent in the data.

\begin{table}[t]
\centering
\caption{Polynomial Kernel SVM for Balanced Dataset}
\label{tab:poly-svm-balanced-results}
\begin{tabular}{ccc}
\toprule
Resolution Size & Accuracy Score & Confusion Matrix \\
\midrule
10   & 0.894737 & $[[167, 23], [17, 173]]$ \\
20   & 0.905263 & $[[170, 20], [16, 174]]$ \\
30   & 0.897368 & $[[167, 23], [16, 174]]$ \\
40   & 0.900000 & $[[168, 22], [16, 174]]$ \\
50   & 0.892105 & $[[166, 24], [17, 173]]$ \\
60   & 0.889474 & $[[166, 24], [18, 172]]$ \\
70   & 0.900000 & $[[169, 21], [17, 173]]$ \\
80   & 0.900000 & $[[169, 21], [17, 173]]$ \\
90   & 0.900000 & $[[169, 21], [17, 173]]$ \\
100  & 0.894737 & $[[168, 22], [18, 172]]$ \\
150  & 0.892105 & $[[167, 23], [18, 172]]$ \\
200  & 0.892105 & $[[167, 23], [18, 172]]$ \\
250  & 0.889474 & $[[166, 24], [18, 172]]$ \\
\bottomrule
\end{tabular}
\end{table}

\begin{table}[t]
\centering
\caption{Polynomial Kernel SVM for Unbalanced Dataset}
\label{tab:poly-svm-unbalanced-results}
\begin{tabular}{ccc}
\toprule
Resolution Size & Accuracy Score & Confusion Matrix \\
\midrule
10   & 0.961369 & $[[0, 0], [35, 871]]$ \\
20   & 0.967991 & $[[0, 0], [29, 877]]$ \\
30   & 0.972406 & $[[0, 0], [25, 881]]$ \\
40   & 0.971302 & $[[0, 0], [26, 880]]$ \\
50   & 0.971302 & $[[0, 0], [26, 880]]$ \\
60   & 0.972406 & $[[0, 0], [25, 881]]$ \\
70   & 0.974614 & $[[0, 0], [23, 883]]$ \\
80   & 0.974614 & $[[0, 0], [23, 883]]$ \\
90   & 0.975717 & $[[0, 0], [22, 884]]$ \\
100  & 0.974614 & $[[0, 0], [23, 883]]$ \\
150  & 0.976821 & $[[0, 0], [21, 885]]$ \\
200  & 0.976821 & $[[0, 0], [21, 885]]$ \\
250  & 0.976821 & $[[0, 0], [21, 885]]$ \\
\bottomrule
\end{tabular}
\end{table}

\begin{table}[t]
\centering
\caption{Gaussian Kernel SVM for Balanced Dataset}
\label{tab:gaussian-svm-balanced-results}
\begin{tabular}{ccc}
\toprule
Resolution Size & Accuracy Score & Confusion Matrix \\
\midrule
10   & 0.907895 & $[[173, 17], [18, 172]]$ \\
20   & 0.902632 & $[[173, 17], [20, 170]]$ \\
30   & 0.902632 & $[[174, 16], [21, 169]]$ \\
40   & 0.905263 & $[[175, 15], [21, 169]]$ \\
50   & 0.910526 & $[[175, 15], [19, 171]]$ \\
60   & 0.913158 & $[[175, 15], [18, 172]]$ \\
70   & 0.910526 & $[[174, 16], [18, 172]]$ \\
80   & 0.907895 & $[[174, 16], [19, 171]]$ \\
90   & 0.907895 & $[[174, 16], [19, 171]]$ \\
100  & 0.907895 & $[[174, 16], [19, 171]]$ \\
150  & 0.910526 & $[[173, 17], [17, 173]]$ \\
200  & 0.905263 & $[[174, 16], [20, 170]]$ \\
250  & 0.918421 & $[[175, 15], [16, 174]]$ \\
\bottomrule
\end{tabular}
\end{table}

\begin{table}[t]
\centering
\caption{Gaussian Kernel SVM for Unbalanced Dataset}
\label{tab:gaussian-svm-unbalanced-results}
\begin{tabular}{ccc}
\toprule
Resolution Size & Accuracy Score & Confusion Matrix \\
\midrule
10   & 0.967991 & $[[0, 0], [29, 877]]$ \\
20   & 0.960265 & $[[0, 0], [36, 870]]$ \\
30   & 0.960265 & $[[0, 0], [36, 870]]$ \\
40   & 0.962472 & $[[0, 0], [34, 872]]$ \\
50   & 0.967991 & $[[0, 0], [29, 877]]$ \\
60   & 0.969095 & $[[0, 0], [28, 878]]$ \\
70   & 0.969095 & $[[0, 0], [28, 878]]$ \\
80   & 0.970199 & $[[0, 0], [27, 879]]$ \\
90   & 0.970199 & $[[0, 0], [27, 879]]$ \\
100  & 0.971302 & $[[0, 0], [26, 880]]$ \\
150  & 0.977925 & $[[0, 0], [20, 886]]$ \\
200  & 0.970199 & $[[0, 0], [27, 879]]$ \\
250  & 0.983444 & $[[0, 0], [15, 891]]$ \\
\bottomrule
\end{tabular}
\end{table}

In short, this detailed analysis has illuminated the varying degrees of success among the classifiers employed, providing critical insights into their relative performance. The Sigmoid Kernel SVM, owing to its subpar performance, is evidently ill-suited for the classification task at hand. Conversely, the Polynomial Kernel SVM has proven to be a more adept choice when compared to logistic regression, thanks to its ability to model complex relationships. Finally, the Gaussian Kernel SVM has emerged as the most proficient classifier, particularly when dealing with datasets replete with intricate and nonlinear patterns, establishing its superiority among the classifiers considered in this assessment. These findings provide valuable guidance for making informed decisions about the choice of classifier in future endeavors, ensuring optimal performance in classification tasks. We can increase the efficiency of the SVM model by using methods as in \cite{Fachrurrozi} but our main focus here is to analyze the performance.

The data presented in Tables 1 to 8 comprises key performance metrics, including Resolution Size, Accuracy Score, and the Confusion Matrix. The Confusion Matrix, a pivotal component of this assessment, is delineated by four fundamental parameters: TP (True Positive), FP (False Positive), TN (True Negative), and FN (False Negative). Here in the tables provided Confusion Matrix is as [[TP, FP], [FN, TN]] Structurally, the confusion matrix is represented as a 2x2 matrix with the following format:

\[
\begin{pmatrix}
TP & FP \\
FN & TN
\end{pmatrix}
\]

In this representation, `TP' signifies the count of True Positives, `FP' indicates the count of False Positives, `TN' denotes the count of True Negatives, and `FN' enumerates the count of False Negatives. These values are essential in gauging the accuracy and efficacy of the classification models under consideration, providing valuable insights into their performance in distinguishing between positive and negative instances within the dataset. The accuracy score for a given confusion matrix is defined as: $$\text{Acuracy Score} = \frac{TP+TN}{TP+FP+FN+TN}$$

We shall now proceed to conduct a detailed examination of performance, specifically in relation to its connection with the resolution size. This investigation will enable us to gain a comprehensive understanding of how various performance metrics are influenced by changes in the resolution size. In Figure 3, we have presented a series of plots illustrating the interplay between Accuracy scores and Resolution Size.

\begin{figure}[h!]
     \centering
     \begin{subfigure}[b]{0.49\textwidth}
         \centering
         \includegraphics[width=\textwidth]{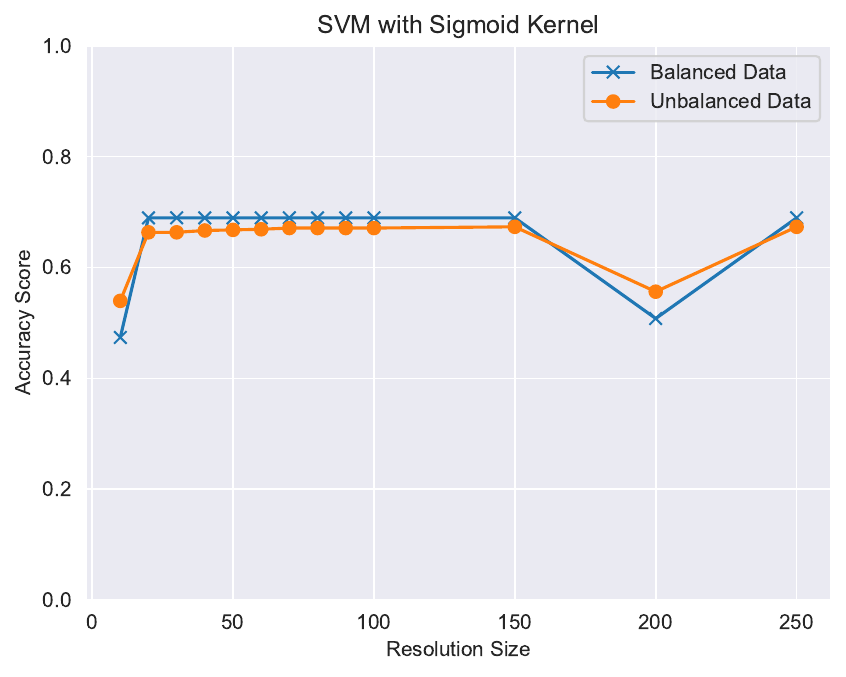}
         \caption{}
         \label{logistic regression on balanced dataset}
     \end{subfigure}
     \hfill
     \begin{subfigure}[b]{0.49\textwidth}
         \centering
         \includegraphics[width=\textwidth]{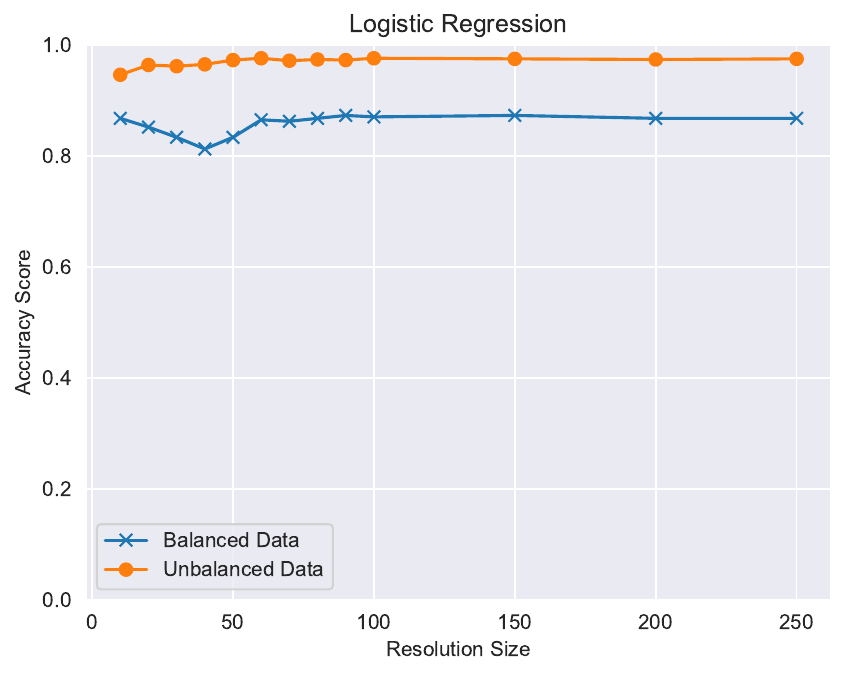}
         \caption{}
         \label{logistic regression on unbalanced dataset}
     \end{subfigure}
     \hfill
     \begin{subfigure}[b]{0.49\textwidth}
         \centering
         \includegraphics[width=\textwidth]{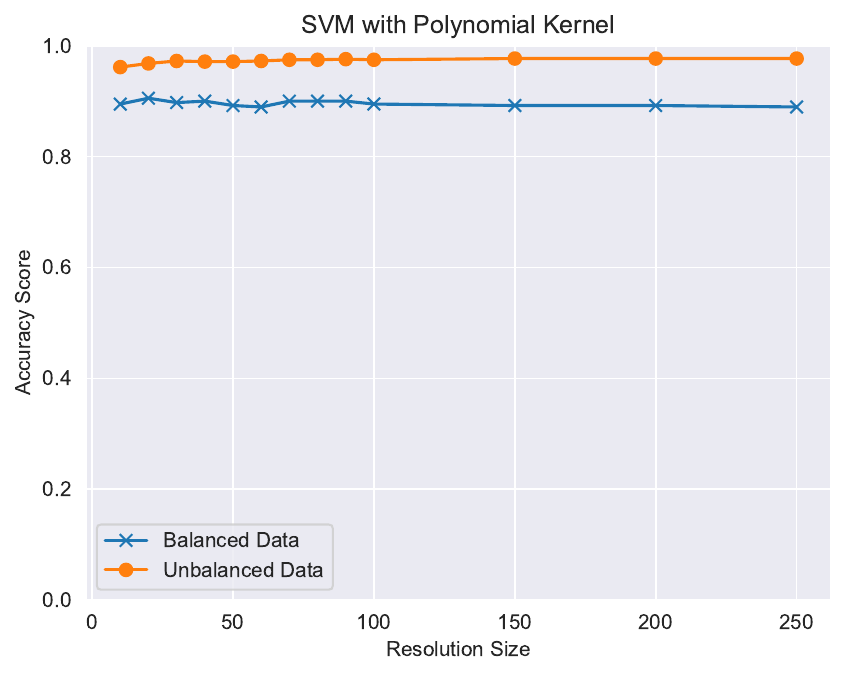}
         \caption{}
         \label{logistic regression on balanced dataset}
     \end{subfigure}
     \hfill
     \begin{subfigure}[b]{0.49\textwidth}
         \centering
         \includegraphics[width=\textwidth]{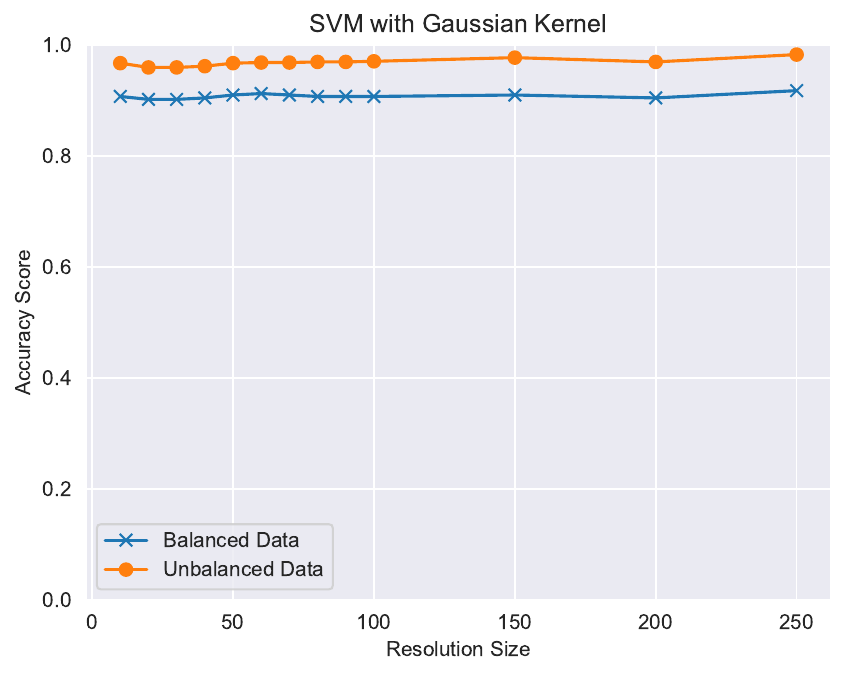}
         \caption{}
         \label{logistic regression on unbalanced dataset}
     \end{subfigure}
     \hfill
     \caption{Plots Representing Accuracy Score (vertical axis) and Resolution Size (horizontal axis) for (a) Logistic Regression, (b) SVM with Sigmoid Kernel, (c) SVM with Polynomial Kernel, (d) SVM with Gaussian Kernel}
\end{figure}

The data indicates a consistent trend of increasing accuracy as the resolution size expands, especially for the Logistic Regression model and SVMs using both Polynomial and Gaussian Kernels. Importantly, this gradual increase in accuracy aligns with larger resolution settings, implying that more detailed data representation leads to improved model performance.

However, an interesting deviation is observed when looking at the SVM with the Sigmoid Kernel. At a resolution size of 200, there is a notable and somewhat perplexing drop in accuracy. This dip in accuracy is, however, rectified when the resolution is set to 250. Such fluctuations in accuracy can be attributed to several factors, often dependent on the specifics of the training and test datasets used. It is crucial to note that the dataset in question, being relatively small, is prone to these changes. While this occurrence is not rare, it poses a considerable challenge in the quest to develop robust classification models. Therefore, the importance of addressing and reducing such accuracy fluctuations, especially at critical resolution settings, is emphasized, as it is crucial for the creation of dependable classification models.

Another significant observation relates to how our models perform when used on unbalanced datasets. In this case, it is clear that the accuracy scores generally maintain a higher average compared to those in the balanced dataset. However, this pattern does not apply to the SVM with the Sigmoid Kernel, which sees a drop in accuracy.

The reason for this occurrence is rooted in the inherent differences between balanced and unbalanced datasets, such as changes in statistical measures like variance, skewness, and other pertinent factors. These differences can significantly affect the performance results of classification models. The ability of the models to handle the inherent imbalances in the context of the unbalanced dataset is crucial in determining their accuracy scores.

\begin{table}[t]
    \centering
    \caption{Logistic Regression on Balanced Dataset}
    \label{tab:logistic_regression_balanced}
    \begin{tabular}{cccc}
        \toprule
        {Resolution Size} & {TPR} & {FPR} & {F1-Score} \\
        \midrule
        10 & 0.868421 & 0.131579 & 0.868421 \\
        20 & 0.848958 & 0.143617 & 0.853403 \\
        30 & 0.832461 & 0.164021 & 0.834646 \\
        40 & 0.818182 & 0.191710 & 0.811671 \\
        50 & 0.846995 & 0.177665 & 0.831099 \\
        60 & 0.883978 & 0.150754 & 0.862534 \\
        70 & 0.863158 & 0.136842 & 0.863158 \\
        80 & 0.876344 & 0.139175 & 0.867021 \\
        90 & 0.873684 & 0.126316 & 0.873684 \\
        100 & 0.877005 & 0.134715 & 0.870027 \\
        150 & 0.873684 & 0.126316 & 0.873684 \\
        200 & 0.872340 & 0.135417 & 0.867725 \\
        250 & 0.876344 & 0.139175 & 0.867021 \\
        \bottomrule
    \end{tabular}
\end{table}

\begin{table}[t]
    \centering
    \caption{Sigmoid Kernel SVM on Balanced Dataset}
    \label{tab:sigmoid_svm_balanced}
    \begin{tabular}{cccc}
        \toprule
        {Resolution Size} & {TPR} & {FPR} & {F1-Score} \\
        \midrule
        10 & 0.470930 & 0.524038 & 0.447514 \\
        20 & 0.669811 & 0.285714 & 0.706468 \\
        30 & 0.669811 & 0.285714 & 0.706468 \\
        40 & 0.669811 & 0.285714 & 0.706468 \\
        50 & 0.669811 & 0.285714 & 0.706468 \\
        60 & 0.669811 & 0.285714 & 0.706468 \\
        70 & 0.671429 & 0.288235 & 0.705000 \\
        80 & 0.671429 & 0.288235 & 0.705000 \\
        90 & 0.671429 & 0.288235 & 0.705000 \\
        100 & 0.671429 & 0.288235 & 0.705000 \\
        150 & 0.671429 & 0.288235 & 0.705000 \\
        200 & 0.508475 & 0.492611 & 0.490463 \\
        250 & 0.671429 & 0.288235 & 0.705000 \\
        \bottomrule
    \end{tabular}
\end{table}

\begin{table}[t]
    \centering
    \caption{Polynomial Kernel SVM for Balanced Dataset}
    \label{tab:poly_svm_balanced}
    \begin{tabular}{cccc}
        \toprule
        {Resolution Size} & {TPR} & {FPR} & {F1-Score} \\
        \midrule
        10 & 0.907609 & 0.117347 & 0.893048 \\
        20 & 0.913978 & 0.103093 & 0.904255 \\
        30 & 0.912568 & 0.116751 & 0.895442 \\
        40 & 0.913043 & 0.112245 & 0.898396 \\
        50 & 0.907104 & 0.121827 & 0.890080 \\
        60 & 0.902174 & 0.122449 & 0.887701 \\
        70 & 0.908602 & 0.108247 & 0.898936 \\
        80 & 0.908602 & 0.108247 & 0.898936 \\
        90 & 0.908602 & 0.108247 & 0.898936 \\
        100 & 0.903226 & 0.113402 & 0.893617 \\
        150 & 0.902703 & 0.117949 & 0.890667 \\
        200 & 0.902703 & 0.117949 & 0.890667 \\
        250 & 0.902174 & 0.122449 & 0.887701 \\
        \bottomrule
    \end{tabular}
\end{table}

\begin{table}[t]
    \centering
    \caption{Gaussian Kernel SVM for Balanced Dataset}
    \label{tab:gaussian_svm_balanced}
    \begin{tabular}{cccc}
        \toprule
        {Resolution Size} & {TPR} & {FPR} & {F1-Score} \\
        \midrule
        10 & 0.905759 & 0.089947 & 0.908136 \\
        20 & 0.896373 & 0.090909 & 0.903394 \\
        30 & 0.892308 & 0.086486 & 0.903896 \\
        40 & 0.892857 & 0.081522 & 0.906736 \\
        50 & 0.902062 & 0.080645 & 0.911458 \\
        60 & 0.906736 & 0.080214 & 0.913838 \\
        70 & 0.906250 & 0.085106 & 0.910995 \\
        80 & 0.901554 & 0.085561 & 0.908616 \\
        90 & 0.901554 & 0.085561 & 0.908616 \\
        100 & 0.901554 & 0.085561 & 0.908616 \\
        150 & 0.910526 & 0.089474 & 0.910526 \\
        200 & 0.896907 & 0.086022 & 0.906250 \\
        250 & 0.916230 & 0.079365 & 0.918635 \\
        \bottomrule
    \end{tabular}
\end{table}

We also conducted an analysis of the True Positive Rate (TPR) and False Positive Rate (FPR) values for both balanced and unbalanced datasets. These values can be obtained from Tables 9 to 12 using the relations $$FPR=\frac{FP}{FP+TN} \text{ and } TPR=\frac{TP}{TP+FN}.$$ Notably, in the case of the unbalanced dataset, we observed that the False Positive Rate (FPR) consistently equated to zero. Consequently, we opted to exclude the unbalanced dataset from further consideration. Subsequently, we generated a Receiver Operating Characteristic (ROC) curve using the acquired TPR and FPR values for the balanced dataset. The ROC curve provides a visual representation of a model's discriminatory performance across various threshold levels. This comprehensive examination of TPR and FPR, coupled with the ROC curve, forms a robust evaluation of the model's classification performance under differing dataset conditions. An example of such an evaluation can be found in \cite{Wang}.

\textbf{True Positive Rate (TPR):} Also known as sensitivity or recall, TPR is the proportion of actual positive instances correctly identified by a classification model. It is calculated as the ratio of true positives to the sum of true positives and false negatives.

\textbf{False Positive Rate (FPR):} FPR measures the proportion of actual negative instances incorrectly classified as positive by a model. It is computed as the ratio of false positives to the sum of false positives and true negatives.

\textbf{Receiver Operating Characteristic (ROC):} ROC is a graphical representation of a model's performance across various discrimination thresholds. It plots the True Positive Rate against the False Positive Rate, providing insights into the trade-off between sensitivity and specificity.

\textbf{Area Under the ROC Curve (AUC):} AUC quantifies the overall performance of a classification model by calculating the area under the ROC curve. It ranges from 0 to 1, with higher values indicating better discriminatory ability. AUC is a common metric for evaluating the effectiveness of binary classification models.

The results for TPR and FPR values are on Tables 9 to 12.

\begin{figure}[h!]
     \centering
     \begin{subfigure}[b]{0.49\textwidth}
         \centering
         \includegraphics[width=\textwidth]{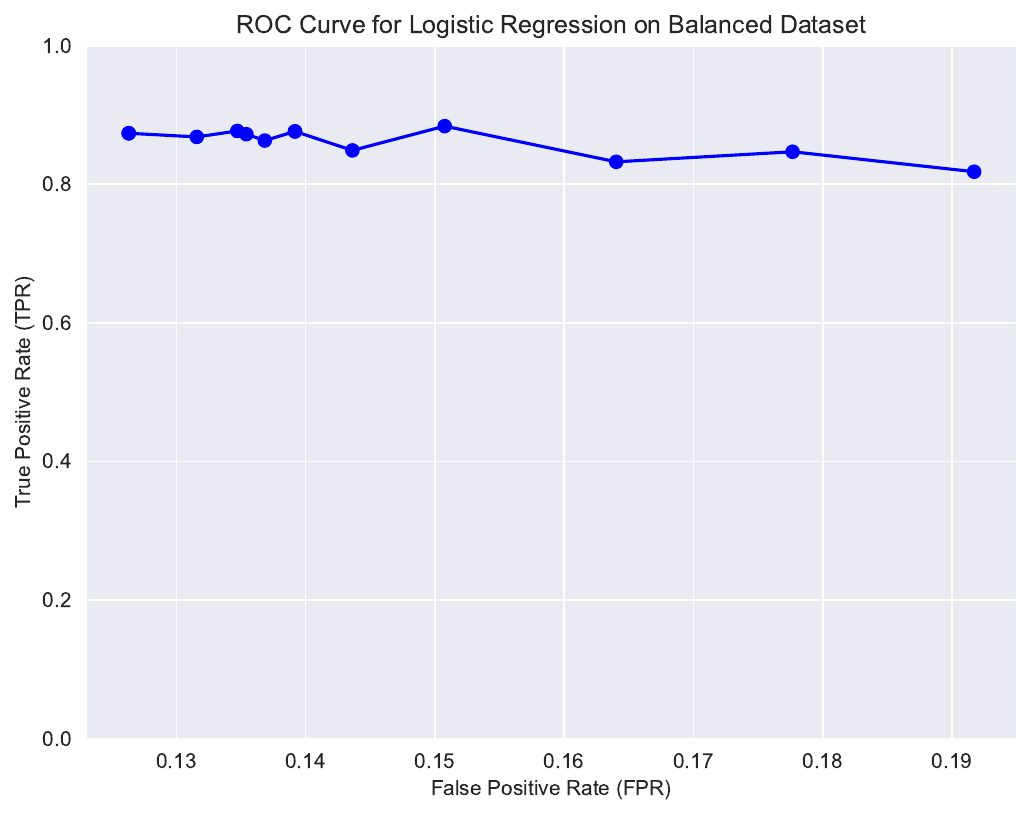}
         \caption{}
         \label{logistic regression on balanced dataset}
     \end{subfigure}
     \hfill
     \begin{subfigure}[b]{0.49\textwidth}
         \centering
         \includegraphics[width=\textwidth]{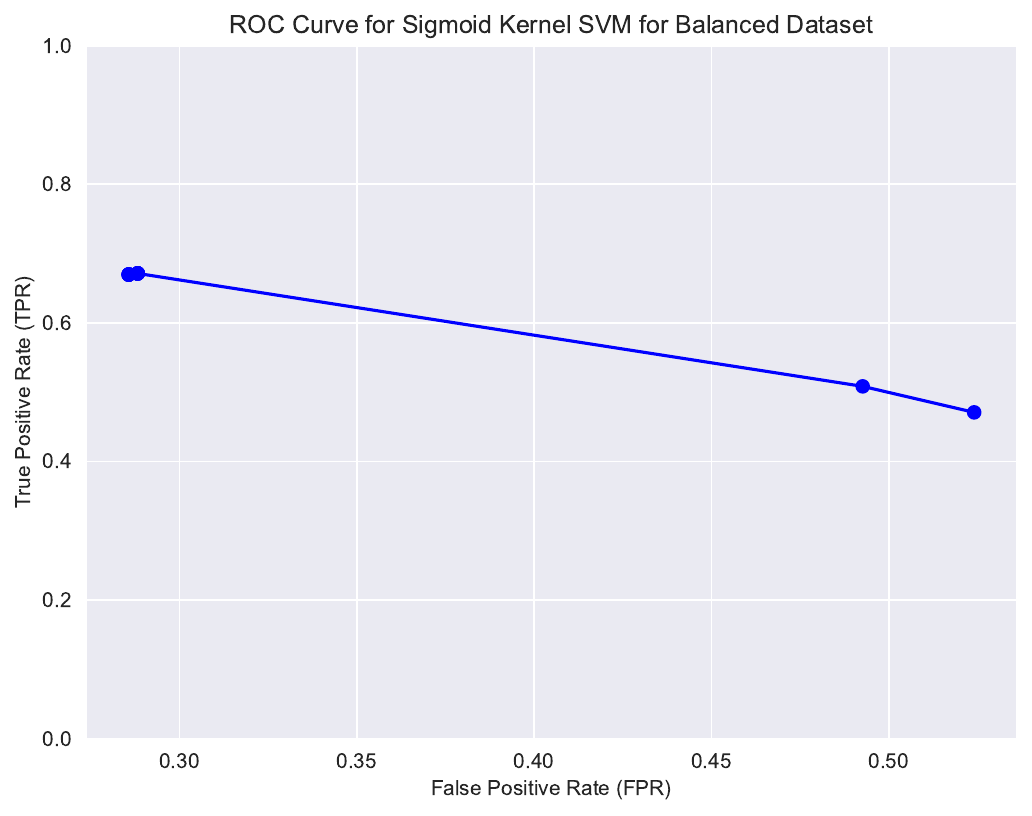}
         \caption{}
         \label{logistic regression on unbalanced dataset}
     \end{subfigure}
     \hfill
     \begin{subfigure}[b]{0.49\textwidth}
         \centering
         \includegraphics[width=\textwidth]{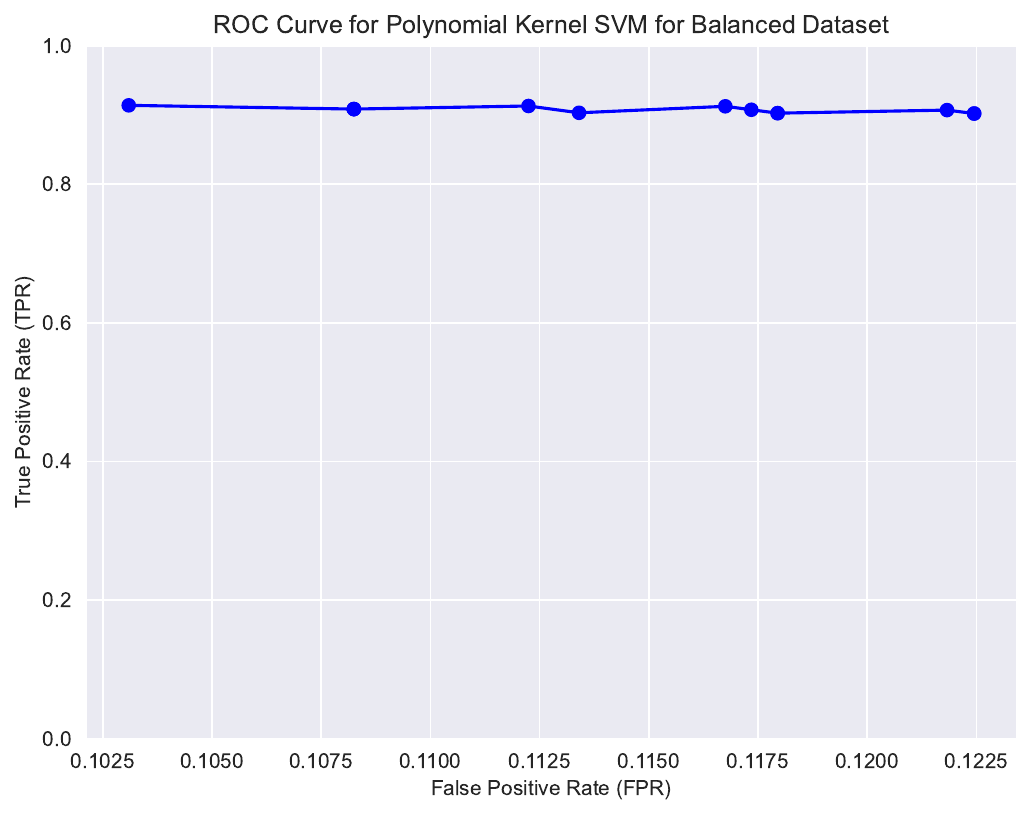}
         \caption{}
         \label{logistic regression on balanced dataset}
     \end{subfigure}
     \hfill
     \begin{subfigure}[b]{0.49\textwidth}
         \centering
         \includegraphics[width=\textwidth]{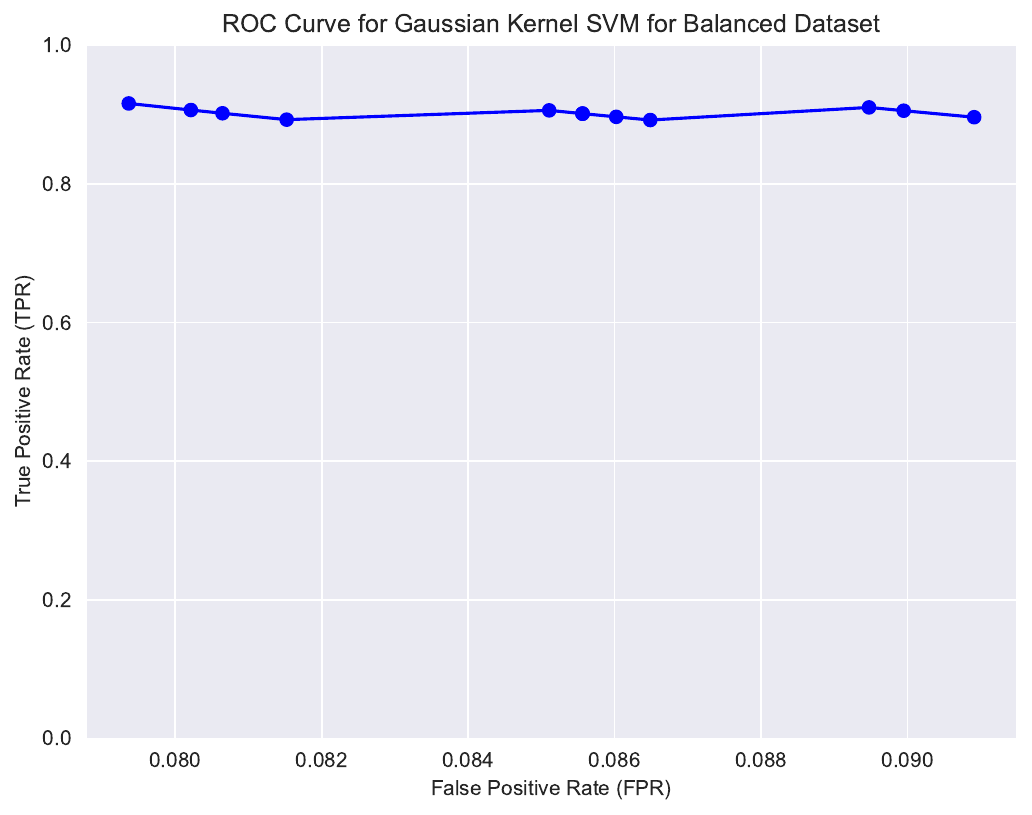}
         \caption{}
         \label{logistic regression on unbalanced dataset}
     \end{subfigure}
     \hfill
     \caption{Plots representing ROC curve on Balanced Dataset for (a) Logistic Regression, (b) SVM with Sigmoid Kernel, (c) SVM with Polynomial Kernel, (d) SVM with Gaussian Kernel}
\end{figure}

We can see the plot of ROC curve for the balanced dataset in Figure 4. We can observe that the curves consistently exhibit a high True Positive Rate (TPR) even as the False Positive Rate (FPR) remains minimal. This performance characteristic indicates the model's robust ability to effectively discriminate between positive and negative instances. The Area Under the ROC Curve (AUC) further supports these findings, with a value close to 1, affirming the model's superior discriminatory power. The details of ROC curves and their significance can be seen in \cite{Krzysztof}. These results collectively highlight the effectiveness of the classification model in striking a robust balance between sensitivity and specificity, thus validating its appropriateness for forest fire detection.

In summary, our findings highlight the critical importance of considering dataset balance and related statistical attributes when developing and evaluating classification models. The close connection between model performance and dataset attributes emphasizes the necessity for a detailed and customized strategy in tackling the complexities of real-world classification tasks.

\section*{Places for Further Improvements}
In the process of developing our Support Vector Machine (SVM) model, several notable challenges have come to the forefront. These challenges can be summarized as follows:

1. \textbf{Resolution Size Impact on Accuracy}: One notable challenge involves the effect of resolution size on accuracy. Even with a considerable information loss due to varying resolution sizes, the corresponding accuracy percentage stayed relatively constant. This dilemma brings up questions about the best resolution size for our model, and how it influences information preservation and classification effectiveness.

2. \textbf{Anomaly in Sigmoid Kernel SVM}: A distinctive anomaly was observed in the performance of the SVM with the Sigmoid kernel. Notably, a sudden drop in accuracy occurred at a resolution size of 200, followed by a subsequent recovery at a resolution of 250. This anomaly highlights the complexities and unpredictabilities in model behavior, thereby giving scope to a deeper understanding of factors influencing such fluctuations.

3. \textbf{Data Set Quantity and Sufficiency}: A challenge faced in our analysis relates to the adequacy of the dataset. It remained inconclusive as to how many instances within the dataset are sufficient for training and assessing the model effectively. Determining the optimal dataset size remains a critical concern, as it directly impacts model generalization and performance.

4. \textbf{Data Augmentation and Pattern Enhancement}: Another unresolved issue surrounds data augmentation, specifically in its capacity to either introduce entirely new patterns or merely enhance existing ones. The distinction between these outcomes is of much importance for evaluating the effectiveness of data augmentation strategies.

5. \textbf{Pixel Attribute Relationships}: Our model is predicated on the attributes of RGB pixel values. Regrettably, we encountered difficulties in drawing meaningful inferences regarding the relationships between color values themselves. This shows that it is challenging to understand how color attributes interact in the classification task.

In summary, these challenges emphasize the complex nature of developing SVM models and analyzing data. Thus, further investigation and deeper understanding of factors such as resolution size, dataset sufficiency, and the impact of model parameters like the Sigmoid kernel, is required.  Addressing these challenges will help in improving the model’s performance and enhancing our understanding of the relationships within the data.

\section*{Significance and Future Works}
Conducting analyses of this nature provides valuable insights into the efficacy of our Support Vector Machine (SVM) model when directly applied to diverse datasets treated as high-dimensional without dimensionality reduction. While our study focused on the forest fire dataset, similar analyses can be extended to other datasets. The results of these evaluations may be used for the creation of algorithms specifically designed for high-dimensional data, taking into account resolution size. This strategy guarantees precision for particular applications while preserving computational speed.

Following the performance analysis of SVM in forest fire detection, future endeavors should focus on refining model efficiency under challenging conditions, particularly with datasets characterized by high dimensionality and limited samples. Further exploration could involve optimizing feature extraction methods, investigating advanced SVM kernel functions, and incorporating ensemble techniques for enhanced predictive accuracy. Additionally, attention should be given to exploring the scalability of the model to larger datasets and evaluating its robustness in diverse environmental contexts.

\section*{Acknowledgments}
We extend our deep appreciation to our colleagues and peers for their significant involvement, inspiring discussions, unwavering encouragement, and collaborative efforts throughout this undertaking. Their diverse viewpoints and contributions have been exceptionally valuable.

We would like to convey our heartfelt gratitude to Mr. Chenna Sai Sandeep and Mr. Suneet Nitin Patil for their constructive input and innovative ideas that greatly improved the implementation and analysis of this project.

We acknowledge that this report would have been notably challenging without the collective commitment and support of all those mentioned above. We sincerely thank everyone for their dedication and contributions, which have transformed this project into a reality.

\section*{Author Contribution} 

\textbf{Ankan Kar}: Conceptualization, Methodology, Validation, Implementation, Visualization, Analysis, Data Collection, Writing – Original Draft, Revised Manuscript.\\
\textbf{Nirjhar Nath}: Validation, Visualization, Implementation, Analysis, Writing -Original Draft, Revised Manuscript.\\
\textbf{Utpalraj Kemprai}: Methodology, Validation, Implementation, Writing – Revised Manuscript.\\
 \textbf{Aman}: Implementation, Writing – Revised Manuscript.

\section*{Conflict of Interest}
The authors declare no conflicts of interest regarding the publication of this paper.

\end{document}